\documentclass[pageno]{jpaper}

\usepackage[normalem]{ulem}
\usepackage{graphicx}
\usepackage{float}
\usepackage[mathletters]{ucs}
\usepackage[utf8]{inputenc}

\begin{document}

\title{Authorship Analysis of Xenophon's Cyropaedia }

\author{Anjalie Field \\ Department of Computer Science, Princeton University \\ Fall Independent Work, 2013 \thanks{This work was advised by Professor Arvind Narayanan (Department of Computer Science, Princeton University) and inspired by Professor Michael Flower (Department of Classics, Princeton University)}}
\date{}

\maketitle

\thispagestyle{empty}

\begin{abstract}
In the past several decades, many authorship attribution studies have used computational methods to determine the authors of disputed texts. Disputed authorship is a common problem in Classics, since little information about ancient documents has survived the centuries. Many scholars have questioned the authenticity of the final chapter of Xenophon's \emph{Cyropaedia}, a 4th century B.C. historical text. In this study, we use N-grams frequency vectors with a cosine similarity function and word frequency vectors with Naive Bayes Classifiers (NBC) and Support Vector Machines (SVM) to analyze the authorship of the \emph{Cyropaedia}. Although the N-gram analysis shows that the epilogue of the \emph{Cyropaedia} differs slightly from the rest of the work, comparing the analysis of Xenophon with analyses of Aristotle and Plato suggests that this difference is not significant. Both NBC and SVM analyses of word frequencies show that the final chapter of the \emph{Cyropaedia} is closely related to the other chapters of the \emph{Cyropaedia}. Therefore, this analysis suggests that the disputed chapter was written by Xenophon. This information can help scholars better understand the \emph{Cyropaedia} and also demonstrates the usefulness of applying modern authorship analysis techniques to classical literature.
\end{abstract}

\section{Introduction}
Statistical techniques have been used to analyze the authorship of texts for over a century, but since Mosteller's and Wallace's famous analysis of the Federalist Papers \cite{wallace64} nearly 50 years ago, computer science has become an essential tool. Authorship attribution is a particularly common problem in classical literature, since authorship controversy surrounds many classical texts. Texts from the Greek and Roman era have barely survived themselves, so any information about their authors can only be gleaned from unreliable secondary sources written centuries later and from the original texts. One such text is Xenophon's \emph{Cyropaedia}. Written in the 4th century B.C, the \emph{Cyropaedia} is an 8-chapter account of the life of Cyrus the Great, the founder of the Persian Empire. The sudden shift in tone from the first 7 chapters to the 8th chapter as well as a number of minor inconsistencies have caused scholars to doubt the authenticity of chapter 8, the epilogue. In general, 19th century scholars believed the epilogue was not authentic. Some suggested it was written by an entirely different author, perhaps even centuries later, or that it was written by Xenophon, but was not part of the original work and was added on later in his life. Most modern scholars believe that the chapter is authentic \cite{hirsch85}.

Whether or not the epilogue is authentic greatly affects how to interpret the rest of the work. Much of the commentary analyzing Xenophon's works focuses on evidence of anti-Persian sentiment,
making the \textit{Cyropaedia} noteworthy, because it provides some of the strongest pro-Persian sentiment in the Xenophontic corpus. Some scholars explain this contrast by claiming that the \emph{Cyropaedia} is entirely a Greek story transported into a Persian setting, meaning the events described are mostly fictional and much of the pro-Persian sentiment can be discarded. In contrast, others believe the work faithfully relates aspects of Persian life, which suggests that Xenophon did have some admiration for Persia. The authenticity of the epilogue greatly impacts whether to interpret the \emph{Cyropaedia} as pro-Persian or anti-Persian. While the rest of the \emph{Cyropaedia} expresses pro-Persian sentiment, the work's epilogue provides strong evidence of anti-Persian sentiment. The epilogue describes the fall of the Persian Empire, specifically its deterioration after Cyrus' death.  Without knowing the authenticity of the \emph{Cyropaedia}'s final chapter, understanding Xenophon's motives in writing the work and using it as a source of information about Persian life is difficult.

Computational techniques for inferring authorship from a text vary widely, though the general approach involves breaking the given text into a set of features and using similarity functions or classifiers to compare the disputed text with other texts. Numerous feature sets have been tried over the past few decades, including word-based analysis, such as vocabulary richness or synonym pairs, syntax-based analysis, such as sentence length, and even meta-data, such as headers/footers and embedded dates \cite{Juola08}.  Some of these feature sets have been rejected as good indicators of authorship, for example word length, but most of the features have been successful in some situations but not in others \cite{holmes94}, generally requiring more analysis to determine the best features to use for authorship studies.

The question of the authenticity of this epilogue differs from a classic authorship attribution problem in a few ways. First, the classic authorship attribution problem consists of a single text of disputed authorship and a number of candidate authors, who may have written the text. The goal of the analysis is to assign the text to the author who mostly likely wrote it by comparing the style of the test text to sample texts written by the candidate authors, called training texts. However, in this case, the chapter has already been attributed to Xenophon. It is nearly impossible to determine a realistic set of candidate authors, since the chapter may have been written at any point over several centuries, and we have no surviving texts for many classical writers. Thus the central question becomes anomaly detection, determining whether or not the epilogue differs stylistically from the other chapters of the \emph{Cyropaedia}, rather than attribution. Additionally, the \emph{Cyropaedia} is written in Ancient Greek. While authorship studies have been conducted in a variety of languages \cite{Juola08}, there has been much more focus on modern languages than on ancient languages. A few studies of Ancient Greek do exist, notably Morton's analysis of Greek prose, which focused on sentence length \cite{morton}, and Ledger's analysis of Plato \cite{ledger89}.

The central purpose of this project was to apply modern stylistic analysis tools to Xenophon's \emph{Cyropaedia} to determine whether or not the epilogue is stylistically similar to the rest of the work. Thus, the study provides evidence on the authorship of the \emph{Cyropaedia}. It additionally demonstrates the application of authorship attribution techniques to an anomaly detection problem. More generally, it shows the usefulness of modern authorship techniques in analyzing classical literature.

The primary features examined we examined were N-grams and word frequencies, and our primary analysis techniques were cosine similarity, Naive Bayesian Classifiers, and Support Vector Machines. The decision to use these methods is discussed in Section 2. Section 3 provides implementation details and Section 4 presents the results. Section 5 discusses these results specifically, while Section 6 discusses related works, and Section 7 concludes. These tests failed to identify significant stylistic differences between the epilogue of the \emph{Cyropaedia} and the other chapters, which suggests that the epilogue was written by Xenophon.

\section{Approach}

\subsection{Overview}
Our general approach was to compare the final chapter of the \emph{Cyropaedia} with the other \emph{Cyropaedia} chapters in order to determine if it is stylistically different. For each feature set and classifier chosen, each chapter of the \emph{Cyropaedia} was individually treated as a test text and the other 6 undisputed chapters (excluding the epilogue) were treated as training texts. In this way, the attribution techniques were adapted to perform anomaly detection. We also used text from other authors for comparison, specifically Plato, Aristotle, and Polybius. The choice of attribution techniques and training texts is discussed below.

\subsection {N-Gram Frequencies}
N-gram frequency analyses are a standard authorship attribution technique. N-gram extraction treats the entire text as a single string, ignoring whitespace. The number of occurrences of each consecutive set of N letters is counted to obtain frequencies (ex. the phrase "I am" contains the 2-grams "Ia" and "am"). One of the primary advantages of N-grams is that they provide context information, meaning N-gram frequencies reflect the author's word choice as well as word combinations.  Other stylistic markers, such as word frequencies and vocabulary richness, provide information about the author's word selection, but not about how the author uses these words in combinations  \cite{Stama09}. N-gram based algorithms are also usually language independent, meaning an algorithm that works successfully with English is also likely to have high accuracy in other languages. Keselj (2003) used a profile-based authorship attribution method that involved creating a profile for each candidate author based on byte-level N-gram frequencies and achieved over 90\% accuracy for classifying Modern Greek Texts, with the best results occurring for $3 \le N \le 5$ \cite{Keselj03}. The language independence of N-grams, as well as their simplicity and accuracy make them a natural choice of feature set in this study. We compared vectors of N-gram frequencies using cosine similarly. Cosine similarity, which measures the the distance between two vectors by the cosine of the angle between them, is a standard distance metric. Its main advantages include simplicity and efficiency \cite{singhal2001modern}.

\subsection{Word Frequencies}
Word frequencies are another common feature set. They rely on the assumption that an author's choice of words is indicative of the author's style, so even across genres, an author will tend to use certain vocabulary. Variations involve changing the set of words examined. Suggested methods include using "synonym pairs", for example, the choice of "big" over "large" \cite{Joa97}, but more common is the use of "function words". Function words include particles, prepositions, and conjunctions, which serve as structural elements in a sentence. One of the problems with word frequencies is content dependency, meaning an author's word choice is influenced by the genre and topic of their writing, so the words may be more indicative of the text's contents than the author's style. Hypothetically, an author would use similar function words regardless of genre, so these words could specifically target an author's style. Generating an objective list of function words to use can be difficult, especially in a language like Ancient Greek, which few authorship studies have focused on. 

Instead, a list of the most common words in an text can serve as an approximate list of function words, since the most common words in any language tend to be structural rather than content-specific. Words like "the", "and", and "that" tend to occur more frequently than words like "education" \cite{Juola08}. Using the 50 most common words has been successful \cite{Burrows89} in distinguishing between authors. Function words are a particularly appropriate metric for analyzing Ancient Greek, because the language uses many particles. In this study, word frequencies were generated for all of the words occurring in the sample texts. We simulated a list of function words by only examining frequencies of the 50, 75, and 100 most common words.

In order to analyze the word frequencies, Multinomial Naive Bayesian Classification (NBC) and Support Vector Machines (SVM) were used. NBC uses conditional probabilities to calculate the probability that a given text belongs to a particular class \cite{Peng04}. The main drawback is that NBC  assumes that features are conditionally independent, which is often not true in texts. However, NBC has been shown to provide accurate classification even with dependent features, though the probabilities generated are not always correct \cite{domingos1996beyond}. NBC is a common classification technique that performs well, even when compared with more complex algorithms.

Support Vector Machines have become widely used in authorship attribution. SVM defines a hyperplane to separate sets of training texts, seeking to minimize true error \cite{Wu04}. They are able to handle high-dimensional data, they eliminate the need for feature selection, and they do not require fine parameter tuning. Additionally, SVMs have been shown to outperform other methods, including Naive Bayesian Classifiers \cite{Joa97}.

\subsection{Comparison Texts}
Three authors were used as comparisons in some of the authorship tests: Plato, Aristotle, and Polybius. Plato's \emph{Laws} serves as a comparison text since it was written around the same time period as the \emph{Cyropaedia} and discusses similar themes. Many scholars even believe that  \emph{Laws} was written as a response to the \emph{Cyropaedia} \cite{hirsch85}. Additionally, the chapters in  \emph{Laws} are close in length to the \emph{Cyropaedia}, specifically in the number of characters. However,  \emph{Laws} differs from the \emph{Cyropaedia} in that it is written as a dialogue, whereas the \emph{Cyropaedia} is plain prose.  Aristotle's \emph{Metaphysics} was chosen for comparison since it was also written around the same time as the \emph{Cyropaedia}. Unlike \emph{Laws}, \emph{Metaphysics} is plain prose, so it can more directly be compared to the \emph{Cyropaedia}. Polybius's \emph{Histories} was written about 200 years after the \emph{Cyropaedia}, but it discusses history, especially focusing on the reasons behind the rise of the Romans. Thus, it discusses some of the same topics as the \emph{Cyropaedia} and is also a historical text, which makes it a reasonable choice for comparison. Additionally, in order to better judge how the chosen feature sets reflect style, comparisons were also made with the chapters of the \emph{Anabasis}, a work attributed to Xenophon without dispute. Using the \emph{Anabasis} chapters attempts to avoid interference from text content, in which the \emph{Cyropaedia} chapters appear similar to each other simply because they all discuss the same subject matter. 

\section{Methods}

\subsection{Texts}
The text of the \emph{Cyropaedia} was obtained from the Perseus Digital Library. It is a digitized version of  Xenophontis opera omni vol. 4. Oxford, Clarendon Press. (1910). The other sample texts used in this study, Aristotle's \emph{Metaphysics}, Plato's \emph{Laws}, Polybius's \emph{Histories}, and Xenophon's \emph{Anabasis} were also obtained from Persues \cite{perseus}.

\subsection{N-grams}
We first extracted N-gram frequencies from the \emph{Cyropaedia}. For each chapter, we counted the frequencies of all N-grams and divided them by the total number of N-grams in the chapter. Then, we created a master list of all N-grams occurring in the work. For each chapter, we created parallel frequency vectors using this master list. 0 was entered as the frequency for any N-gram not occurring in the chapter, so that each chapter was represented by a sparse list of frequencies of N-grams in the master list. We normalized the chapters for text length, by cutting off the end of longer chapters, so that all of the chapters were the same length as the shortest chapter. Each Unicode character in the text was considered a separate character, meaning identical letters with different accents or breathings were considered as different letters. The N-gram vectors were generated using Python scripts.

\subsection{Cosine Similarity}
We used  cosine similarity function to analyze the N-grams vectors. This function was implemented using the \textit{pdist} function in Matlab.

\subsection{Word Frequencies}
Word frequency statistics were downloaded from the Perseus Digital Library. Many Greek forms are ambiguous, meaning different words can have identical forms. For each word, The Perseus Vocabulary Tool determines a maximum frequency, attributing all ambiguous forms that appear in the text to the word, a minimum frequency, attributing no ambiguous forms to the word, and a weighted frequency. The weighted frequency assigns a weight to each inflected form according to how many dictionary forms the inflected word could be attributed to \cite{perseus}. In this study, the weighted frequency per 10,000 words was used to approximate true frequencies. A master list of words was created, containing all of the words occurring in the training and test text and their combined frequencies. For the "All Words" studies, each chapter was represented using all of the words in the master list, meaning a frequency vector was created for each chapter, which contained a value for each word in the master list. For the "Most Common N" studies, a list of the most common N words was generated from the master list. Each chapter was represented as a frequency vector of length N, containing the frequencies of the words in the N Most Common list. The frequency vectors were generated using Python.

\subsection{Classifiers}
Multinomial Naive Bayesian Classification was implemented using the sklearn.naive\_bayes package from scikit-learn. SVM was also implemented using sklearn package from scikit-learn \cite{scikit-learn}. A linear kernel was used (i.e. svm.SVC( kernel="linear")). Most text classification problems are linearly separable \cite{Joa97}, and more complex kernel functions can be slower and often do not increase accuracy. In particular, RBF functions can be sensitive to parameter selection \cite{Diederich03}. Before using SVM classification, the frequency vectors were scaled using preprocessing.MinMaxScaler() from scikit-learn.

\section{Results}

Figures 1-4 show the results of the N-gram frequency analysis. The comparisons between just the \emph{Cyropaedia} chapters, using frequencies of 2-grams, 3-grams, and 4-grams are shown in Figure 1. As the value of N increases, the distances increase overall. In the 2-gram frequencies, all of the chapters are approximately the same distance from each other. In the 3-gram frequency, the first chapter is noticeably further away from the other chapters than they are from each other, and in the 4-gram frequencies, the epilogue is also slightly more distant from the other chapters.

\begin{figure}[H]
  \centering
    \includegraphics[width=0.5\textwidth]{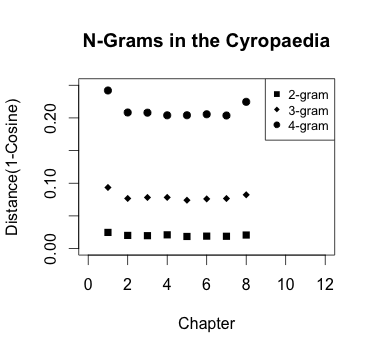}
  \caption{Each point in this figure represents the average distance between the specified chapter and every other chapter in the \emph{Cyropaedia}, excluding the epilogue. The rightmost point, chapter 8, represents the average distance between the epilogue and every other chapter in the \emph{Cyropaedia}. The distance was calculated by representing each chapter as an N-gram frequency vector, either of 2, 3, or 4 grams, and calculating 1 - cosine of the angle between each pair of vectors. These distances were averaged to obtain a single value for each chapter.}
\end{figure}

Figure 2 performs the same analysis as Figure 1, but uses Plato's \emph{Laws} instead of Xenophon's \emph{Cyropaedia}. The same data from Figure 1, 4-gram frequencies in the \emph{Cyropaedia}, is also displayed in this figure in order to compare the spread between chapters. As in Figure 1, the distances between chapters increases with N, so that the chapters are all approximately the same distance apart for 2-grams, but more scattered for 4-grams. The comparison between the \emph{Cyropaedia} and the \emph{Laws} demonstrates that although chapters 1 and 8 of \emph{Cyropaedia} separate slightly from the other chapters in the 4-gram frequencies, the chapters of \emph{Laws}, especially 5 and 6, separate even more so.

\begin{figure}[H]
  \centering
    \includegraphics[width=0.5\textwidth]{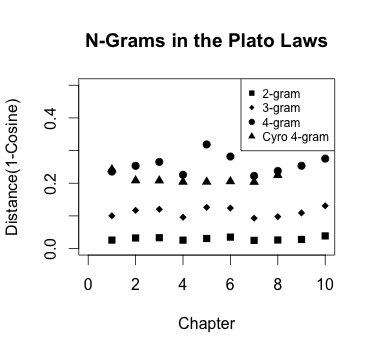}
  \caption{Each point in this figure represents the average distance between the specified chapter and every other chapter in Plato's Laws, excluding the last chapter. The distance was calculated by representing each chapter as an N-gram frequency vector, either of 2, 3, or 4 grams, and calculating 1 - cosine of the angle between each pair of vectors. The triangles, labeled "4-gram Cyro", display the same data shown in the previous figure, the average distances between the \emph{Cyropaedia} chapters, when represented as vectors of 4-grams.}
\end{figure}

Figure 3 demonstrates an N-gram frequency analysis, where the last chapter of the study was written by a different author from the other chapters, specifically the \emph{Cyropaedia} chapters are compared with the first chapter of Aristotle's \emph{Metaphysics}. This figure shows a drastic difference between the Aristotle chapter and the Xenophon chapters. In the 3-grams, the cosine similarity distance of Aristotle's chapter is 0.2208, and in the 4-grams, it is  0.4347. The Aristotle chapter is much further from the other 7 chapters than they are from each other.

\begin{figure}[H]
  \centering
    \includegraphics[width=0.5\textwidth]{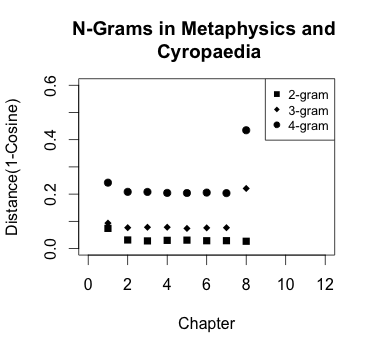}
  \caption{The first 7 points in this figure represent the average distance between the specified chapter and every other chapter in the \emph{Cyropaedia}, excluding the last chapter. The last chapter (chapter 8) of the \emph{Cyropaedia} was replaced with chapter 1 of Aristotle's \emph{Metaphysics}, so chapter 8 represents the average distance between \emph{Metaphysics} chapter 1 and each chapter of the \emph{Cyropaedia}. The distance was calculated by representing each chapter as an N-gram frequency vector, either of 2, 3, or 4 grams, and calculating 1 - cosine of the angle between each pair of vectors.}
\end{figure}

Figure 4 compares the 7 undisputed \emph{Cyropaedia} chapters to chapter 1 of Xenophon's \emph{Anabasis}. The \emph{Anabasis} chapter (represented as chapter 8), shows a clear distance from the other chapters in the 3-grams (0.1271) and 4-grams (0.3101), as well as a slight distance in the 2-grams. The distance is less large than the distances between the \emph{Metaphysics} and the \emph{Cyropaedia} in Figure 3, but the \emph{Anabasis} chapter is still notably different from the \emph{Cyropaedia}.

\begin{figure}[H]
  \centering
    \includegraphics[width=0.5\textwidth]{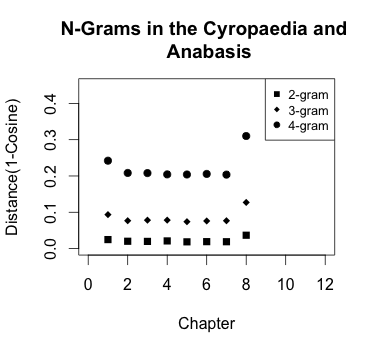}
  \caption{The first 7 points in this figure represent the average distance between the specified chapter and every other chapter in the \emph{Cyropaedia}, excluding the last chapter. The last chapter (chapter 8) of the \emph{Cyropaedia} was replaced with chapter 1 of Xenophon's \emph{Anabasis}, so chapter 8 represents the average distance between \emph{Anabasis} chapter 1 and each chapter of the \emph{Cyropaedia}. The distance was calculated by representing each chapter as an N-gram frequency vector, either of 2, 3, or 4 grams, and calculating 1 - cosine of the angle between each pair of vectors.}
\end{figure}

Figure 5 uses NBC classification to compare each chapter of the  \emph{Cyropaedia} with the works of other authors. A low log probability indicates that the chapter is more closely associated with Xenophon than with the other author (Polybius or Aristotle). The probabilities are lowest when all words in the texts are used as the feature set (i.e. the triangles) and highest when the feature set is only the 50 most common words. This indicates that the \emph{Cyropaedia} chapters are more closely associated with each other when more words are used to generate frequency vectors. Overall, the probability that each chapter is associated with Aristotle or Polybius over Xenophon is about the same for all chapters. Chapters 1, 5, and 8 associated more closely with Xenophon than the other chapters. Since some of the chapters of the \emph{Metaphysics} are very short and the work has many chapters, only a selection were used to represent Aristotle's style, specifically chapters 1-7.

\begin{figure}[H]
  \centering
\hfill
{\includegraphics[width=0.47\textwidth]{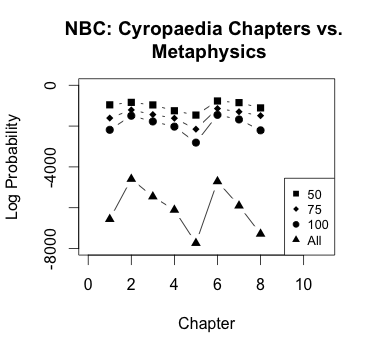}}
\hfill
{\includegraphics[width=0.47\textwidth]{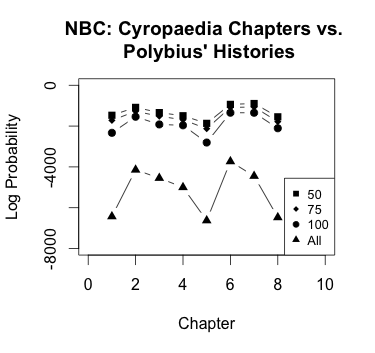}}
\hfill
  \caption{In the left figure, each of the \emph{Cyropaedia} chapters and the \emph{Metaphysics} chapters were represented as a vector of word frequencies. One at at time, each chapter in the \emph{Cyropaedia} was treated as the test text, while the other \emph{Cyropaedia} chapters and the \emph{Metaphyics} chapters were treated as the training texts. A Naive Bayesian Classifier was used to classify the test text as either part of the \emph{Cyropaedia} or part of the \emph{Metaphysics}. All of the \emph{Cyropaedia} chapters were successfully classified as part of the \emph{Cyropaedia}. The probabilities shown are the log of the probability that the given chapter should be grouped in the \emph{Metaphyics}. In the legend, "All" signifies that the frequencies of all of the words occurring in the training texts and test text were used to generate feature vectors. "50", "75", and "100" indicate that the 50, 75, and 100 most common words across the training and test texts were used to generated the feature vectors. In the right figure the exact same algorithm was performed, using Polybius's \emph{Histories} as the second set of training texts instead of the \emph{Metaphysics}.}
\end{figure}

The SVM classification in Figure 6 reveals similar results to the NBC classification in Figure 5. Unlike Figure 5, in Figure 6, a high score indicates a close association with Xenophon. Thus, chapters 1, 5, and 8 show the closest associations with the other \emph{Cyropaedia} chapters. As in Figure 5, the chapters are generally more closely assigned to the \emph{Cyropaedia} when all of the word frequencies in the texts are considers as features.

\begin{figure}[H]
  \centering
  
  \hfill
{\includegraphics[width=0.47\textwidth]{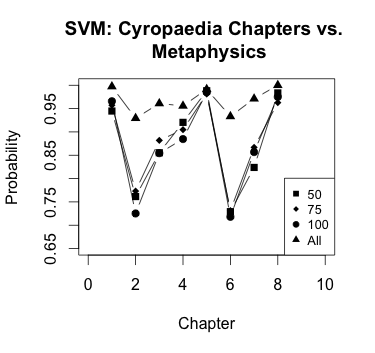}}
\hfill
{\includegraphics[width=0.47\textwidth]{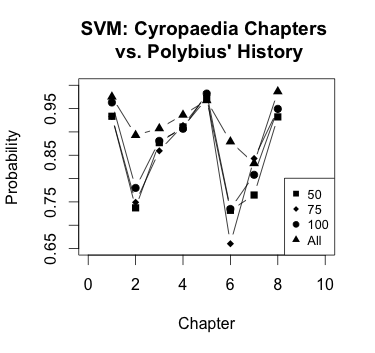}}
\hfill
  \caption{This figure demonstrates the results of the same algorithm described in Figure 5, where each of the \emph{Cyropaedia} chapters was treated as a test text, and the remaining chapters, along with chapters from Aristotle's \emph{Metaphysics} or Polybius's \emph{Histories} were used as training texts. Here SVM was used to classify the given chapter. All chapters were successfully attributed to the \emph{Cyropaedia}. The results displayed are the probability that the given chapter is a part of the \emph{Cyropaedia}. As in the previous figure, the legend distinguishes whether all words, the 50, 75, or 100 most common words were used as the feature set.}
\end{figure}

Figure 7 attempts to distinguish the \emph{Cyropaedia} chapters by classifying them as either Xenophon or Aristotle. However, instead of using the other \emph{Cyropaedia} chapters as training texts, the chapters of Xenophon's \emph{Anabasis} were used. The Naive Bayesian Classifier attributed all of the \emph{Cyropaedia} chapters to Xenophon rather than Aristotle or Polybius. As in the previous figures, chapters 1, 5, and 8 were attributed to Xenophon with the highest probability. Additionally, using all of the word frequencies attributed the chapters to Xenophon with higher probabilities than the more selective feature sets.

\begin{figure}[H]
  \centering
  
\hfill
{\includegraphics[width=0.47\textwidth]{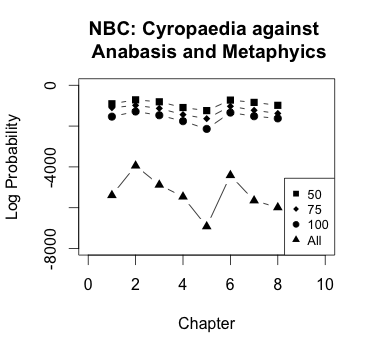}}
\hfill
{\includegraphics[width=0.47\textwidth]{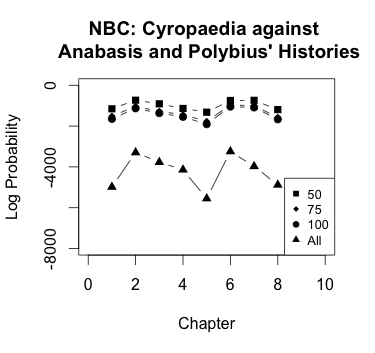}}
\hfill

  \caption{In the left figure, the chapters of Aristotle's \emph{Metaphysics} and Xenophon's \emph{Anabasis} were used as training texts. An NBC classifier was then used to classify each chapter of the \emph{Cyropaedia} as written by Aristotle or Xenophon. All of the \emph{Cyropaedia} chapters were successfully classified as written by Xenophon, and the log probabilities shown are the log of the probability that the given chapter was written by Aristotle. In the right figure, the exact same algorithm was performed, using the \emph{Anabasis} and Polybius's \emph{Histories} as training texts. All of the \emph{Cyropaedia} chapters were successfully classified as Xenophon.  As in the previous figure, the legend identifies whether all words, or the 50, 75, or 100 most common words were used to generated frequency vectors.}
\end{figure}

Figure 8 similarly shows the results of using Xenophon's \emph{Anabasis} as a training text, with an SVM instead of an NBC. Once again, all of the \emph{Cyropaedia} chapters were successfully attributed to Xenophon. As in the previous figure, chapters 1, 5, and 8 were attributed to Xenophon with the highest probabilities, as opposed to Aristotle or Polybius. Additionally, using all of the word frequencies generated higher probabilities than a limited selection.

\begin{figure}[H]
  \centering
 
 \hfill
{\includegraphics[width=0.47\textwidth]{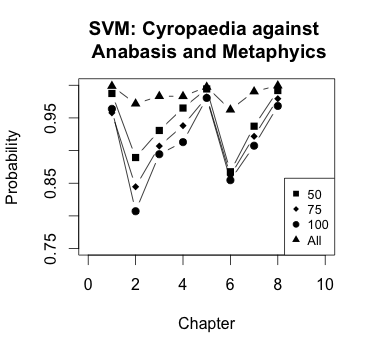}}
\hfill
{\includegraphics[width=0.47\textwidth]{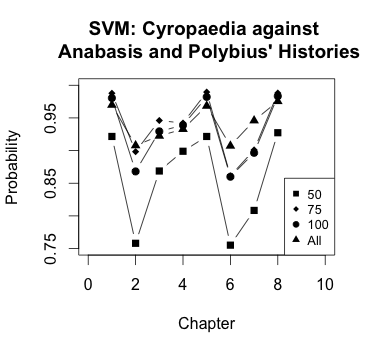}}
\hfill 
  \caption{The same algorithm as in Figure 7 was performed, using the \emph{Metaphysics} and the \emph{Anabasis} or the \emph{Histories} and the \emph{Anabasis} as training texts. Here, an SVM was used instead of a NBC to classify the \emph{Cyropaedia} chapters. The probabilities shown are the probability that the given chapter was written by Xenophon, as opposed to Aristotle or Polybius. As in the previous figure, the legend distinguishes whether all words, the 50, 75, or 100 most common words were used as the feature set.}
\end{figure}

\section{Discussion}

Figure 3 supports the validity of using N-grams to distinguish between authors. In the 2-grams, all of the chapters tested are close together. However, in the 3-grams, the \emph{Cyropaedia} vectors are all close to each other ($< .1$), while the Aristotle chapter is much further away from the \emph{Cyropaedia} chapters ($>.2$). Similarly in the 4-grams, chapter 1 of Aristotle's \emph{Metaphysics} is significantly further away from the \emph{Cyropaedia} chapters than they are from each other (.4 vs .2). Thus, these results successfully distinguish chapter 1 of Aristotle's \emph{Metaphysics} from the undisputed chapters of the \emph{Cyropaedia}.

However, while the \emph{Cyropaedia} discusses the life of Cyrus, the \emph{Metaphysics} discusses the philosophy of existence, so these two texts describe completely different subject matters. One of the problems with N-grams are their tendency to be too content-dependent \cite{Gamon04}. While N-grams do accent some stylistic markers, like word order, they can also reflect the general vocabulary of a text, which is greatly dependent on the subject matter. Figure 4 demonstrates the problem of content dependency. Although chapter 1 of the \emph{Anabasis} was undisputedly written by Xenophon, the chapter is clearly further away from the \emph{Cyropaedia} chapters than they are from each other. However, the difference is less pronounced than the comparison in Figure 3, between the \emph{Cyropaedia} and Aristotle. In the 3-gram vectors, the \emph{Anabasis} chapter measures a cosine distance of 0.1271, in contrast with Aristotle's 0.2208. In the 4-gram vectors, the \emph{Anabasis} chapters yields an average distance of 0.3101, while the Aristotle chapter is 0.4347. The distance between the two Xenophon works suggests that the N-gram analysis is subject to interference by markers other than authorship, most likely content. However, although the sample size in this study is too small to draw firm conclusions, the smaller distances between the Xenophon texts than between Xenophon and Aristotle suggest that these N-grams do reflect some stylistic measures of authorship.

In the context of Figures 3 and 4, the results displayed in Figure 1 suggest that the epilogue of the \emph{Cyropaedia} is not significantly different from the rest of the work. The 2 and 3-gram vectors show no significant difference between the epilogue and the other chapters of the \emph{Cyropaedia}. In fact, in the 3-gram vectors, it is the 1st chapter, not the 8th chapter, that is slightly further from the other chapters. In the 4-gram vectors, the 1st chapter is again slightly further from the other chapters. Although the 4-gram vectors show that the 8th chapter and the 1st chapter are both slightly further from the other chapters than they are from each other, neither the 1st chapter nor the 8th chapter are nearly as far from the other chapters as the \emph{Metaphyiscs} chapter is in Figure 3 and the \emph{Anabasis} chapter is in Figure 4.

It is possible that the 1st and 8th chapters show slight deviations because of content dependencies. All 8 chapters of the \emph{Cyropaedia} are about Persia and Cyrus the Great, but the 1st chapter serves an introduction. It describes Cyrus's family, appearance, personality and education. The middle 6 chapters (2-7) focus on Cyrus's rise to power and include less relevant material like tactics and sophistic debates. The 8th chapter serves as an epilogue, discussing how Cyrus ruled his empire and the fall of Persia after his death \cite{Gera}. The difference in focus, rather than a difference in authorship, could explain the slight distance between the 1st chapter, the 8th chapter, and the other chapters.

Figure 2 also suggests that some variations within a work are expected. The distances between the 2, 3, and 4-gram vectors are shown for Plato's \emph{Laws}, a work written in the same time period as the \emph{Cyropaedia} and discussing many similar themes. While the 2-gram vectors are all about the same distance apart, the distances between chapters vary more in the 3-gram vectors and even more in the 4-gram vectors. Chapters 5 and 6 in particular are further from the other chapters, but there is much variation overall in the distance each chapter is from the others. In comparison, the 4-gram vectors of the \emph{Cyropaedeia}, also shown in this figure, are all about the same distance apart. The slight deviations of the 1st and 8th chapter are very small when compared to the varying distances in the Plato chapters. This figure suggests that some deviations in the distances between 4-gram vectors are normal, meaning they do not necessarily indicate a significant difference between these chapters and the others.

Therefore, while the 4-gram vectors do suggest that the 8th chapter of the \emph{Cyropaedia} differs slightly from the other chapters, this difference is not enough to conclude the chapter was written by someone other than Xenophon. Especially considering the 1st chapter, whose authorship has not been previously questioned, also differed slightly from the other chapters, it is likely that the slight deviation of the 8th chapter is due to the content-dependency problem of N-grams or the natural variations between chapters of the same work.

While the N-gram analysis compared the \emph{Cyropaedia} chapters directly to each other, the results in Figure 5 and 6 compare the chapters first to Aristotle and Polybius. The chapters are then only compared to each other by their differences from Aristotle and Polybius. Since the \emph{Metaphysics} and the \emph{Histories} are both on different subjects than the \emph{Cyropaedia}, this method allows content dependency to become less important. All of the \emph{Cyropaedia} chapters differ in content from the texts they are being compared to.

The word frequency data, both with Naive Bayesian Classifiers and Support Vector Machines, reveal similar results. Figures 5 and 6 identify \emph{Cyropaedia} chapters 1, 5, and 8 as the chapters attributed to Xenophon with the highest probability. Both of these figures support the authenticity of chapter 8. If chapter 8 were not written by Xenophon, the other 7 chapters would be expected to be attributed to Xenophon with higher probabilities than chapter 8.

Nevertheless, some of the \emph{Cyropaedia} chapters might appear more similar to each other because some of them have more similar content than others. In this study, the use of the N most common words was used to imitate a list of function words, which can reduce the effects of content dependency. In Figures 5 and 6, using the 50, 75, and 100 most common words results in greater distinction between the \emph{Cyropaedia} chapters than using all of the words in the texts. It is possible that the using a smaller feature set reveals Xenophon's style more clearly. When using the full set of words, the content-dependency of some of the words causes the \emph{Cyropaedia} chapters to be associated with each other with very high probabilities, but when using a smaller set of words, the stylistic differences between the chapters become more evident. However, in text classification, there are very few irrelevant features, and even features contributing less information about the authorship of a text can still provide considerable insight \cite{Joa97}. Thus, it is also possible that reducing the feature set causes relevant authorship information to be removed, so the chapters are associated with Xenophon with lower probabilities. The results of reducing the size of the frequency vectors is unclear from the data in this study.

While content dependency may still be an issue in Figures 5 and 6, Figures 7 and 8 attempt to minimize this problem even further by using the \emph{Anabasis} instead of the \emph{Cyropaedia} as training text. Instead of grouping each chapter with other chapters of the \emph{Cyropeadia}, these figures group each chapter of the \emph{Cyropaedia} with the \emph{Anabasis}, or Xenophon more generally. The \emph{Anabasis} relates the story of Greek mercenaries stranded in enemy territory and their march to the Black Sea \cite{Wencis77}. The text does have some overlap with the \emph{Cyropaedia} since the Greek soldiers are initially hired by Cyrus the Younger and are stranded in Persia. However, the \emph{Anabasis} focuses on the story of the Greeks, not Persian campaigns and customs, so much of the content is different from the \emph{Cyropaedia}. In theory, chapters of the \emph{Cyropaedia} are associated with the \emph{Anabasis} because of stylistic similarities instead of content.

Despite the different training texts, Figures 7 and 8 closely resemble Figures 5 and 6. Once again, both figures attribute chapters 1, 5, and 8 to Xenophon with the highest probabilities. Reducing the number of words in the feature set causes the chapters to become more scattered and the probabilities of their attribution to Xenophon to become lower. The similarities between these two different methods could suggest that content dependency is not skewing the authorship classification in Figures 5 and 6, though it is still possible that genre is causing the chapters to be attributed to Xenophon with high probabilities.

One problem with the set up of this study is the choice of training texts. Aristotle's \emph{Metaphysics} and Polybius's \emph{Histories} are both formatted similarly to the \emph{Cyropaedia} in that they are plain prose divided into chapters of about the same lengths. However, Xenophon primarily wrote history, often with didactic or fictional elements, while Aristotle wrote philosophy. Polybius did also write history, but he wrote at a slightly later time period than Xenophon, which could result in stylistic differences. It is difficult to determine how the difference between texts might affect classification probabilities. In Figures 5, 6, 7, and 8, both the Aristotle comparisons and the Polybius comparisons have similar shapes. They both identify chapters 1, 5, and 8 as having the highest probability of attribution to Xenophon. More generally, they follow similar trajectories, across the 50, 75, and 100 most common words, as well as all words. The similarity of these results suggests that the deviations between chapters are reflective of the actual stylistic differences between \emph{Cyropaedia} chapters, rather than misrepresentative probabilities because of the training text selection. However, using only Aristotle and Polybius does not provide a robust comparison and comparisons with more authors could be helpful to ensure the accuracy of the classifiers. Nevertheless, these results again support the authenticity of chapter 8. Across all data sets, it differs little from the other \emph{Cyropaedia} chapters and is attributed to Xenophon with a high probability.

Although neither the N-gram analysis nor the word-frequency tests indicate that the epilogue of the \emph{Cyropaedia} is unauthentic, the N-grams and word frequencies do not distinguish the chapters in the same way. Figure 1 identifies chapters 1 and 8 as the chapters furthest away from the other chapters in the work. In contrast, Figures 5 and 6 indicate that chapters 1, 5, and 8 were attributed to Xenophon with the highest probability, meaning they were most similar to the other chapters of the \emph{Cyropaedia}. Although Figures 7 and 8 compare the chapters to the \emph{Anabasis}, they also indicate that chapters 1, 5, and 8 are attributed to Xenophon with the highest probabilities. Since N-grams are reflective of the word choice in a text, the reversal between the N-grams and the word frequencies is surprising. The difference could be the result of the different classifiers used to analyze the frequency vectors. Cosine similarity calculates a simple distance, while NBC and SVM consider each element in the feature vector within the larger context of the work. Additionally, content dependency may have been a greater problem in N-gram frequencies. Since the differences between chapters overall were small, the differences between the N-grams and word frequencies are not very significant, and it is difficult to fully explain them.

\section{Related works}
Although many scholars have speculated on the authorship of the \emph{Cyropaedia}, no previous studies have used a computational method to analyze the authorship of this text. A few other authorship studies have been conducted on Greek prose with different methods. In particular, A.Q. Morton has conducted multiple studies on Ancient Greek authors. In 1965, he suggested sentence length distributions as a way to profile authors \cite{morton}. While average sentence length has largely been rejected as a measure of authorship \cite{Juola08}, sentence length distributions show more potential \cite{holmes94}. However, this method has a number of drawbacks. First, sentence length is easy for an author to control consciously, so a forger could easily imitate another author's sentence lengths. More importantly, sentence length depends on the punctuation of the text, so it is only a reliable measure when the text uses the original author's punctuation or when a single editor has punctuated all of the texts being compared \cite{holmes94}. It is difficult for classical texts to meet these criteria, since over the past millennia, the texts have changed hands many times and have often been recopied. Most punctuation in Greek texts is entirely modern and not necessarily reflective of the original author's intentions.

Morton also contributed to a study which analyzed the authorship of Plato's Seventh Letter \cite{levison68}. This study also looked at sentence distributions, as well as the distributions of specific particles within sentences. The method in this study has the same drawbacks as the previous one in that it depends on punctuation, which may not be reliable for many texts. The study also only looked at the distributions of 2 particles, while other studies have shown that tests generally perform better when they examine as many features as possible. A different technique involving word position and context had some success in Greek, its lack of success on Elizabethan dramas diminishes its credibility \cite{Smith}. Considering the varied results of prior techniques used on Greek texts, there is a need for tests that can be reliably used. The use of more recent methods in this analysis of the \emph{Cyropaedia} suggests that these modern methods have the potential to be adapted for Greek texts. However, since this project focused on the authorship of a particular work, it did not involve a thorough test of these methods across a variety of texts, so more research is needed to establish the accuracy of N-grams and word frequencies tests with Greek.

A 1982 study of the Corpus Lysiacum did use word frequencies as one of the features to distinguish texts attributed to Lysias. This method used chi-squared tests to provide a distance measure. Chi-squared tests assume that features are independent, but independence is often not true in grammar. The study of the Corpus Lysiacum additionally involved parsing the text by hand, which is extremely time consuming and error prone \cite{usher82}.

A third technique published in 1989 uses percentages of words containing specific letters, having a specific letter as the second-to-last letter in the word, and ending in specific letters as features for a total of 37 variables  \cite{ledgerRev}. While this technique worked for analyzing Plato, it is not apparent why these particular features are good indicators of style. Without justification, it is difficult to apply this method to other texts, especially texts in other languages \cite{holmes94}. Nevertheless, this method does appear to work well for Ancient Greek.

Although these previous works have examined the authorship of Ancient Greek texts, they have not used modern tools now available, such as the electronic accessibility of many texts through databases like Perseus. N-grams and word frequencies have worked successfully in other languages, as have NBC and SVM. These methods have the potential to classify texts or detect outlying chapters with higher accuracy and more flexibility than tests previously tried. Further research should involve trying these methods on a variety of texts of known authorship in order to determine their accuracy.

\section{Conclusions}
The results of the N-grams and word frequencies suggest that the epilogue of the \emph{Cyropaedia} was genuinely written by Xenophon, though these data are not sufficient to conclude if the chapter was part of the original work or added on later in Xenophon's life. Although the N-grams do suggest chapter 8 is further from the other chapters than they are from each other, chapter 1 also appears slightly further from the other chapters. Neither chapter is as far away as chapter 1 of Aristotle's \emph{Metaphyics}, nor a chapter from the \emph{Anabasis}. It is likely that these deviations are due to content dependency or natural variations in an author's style across a long work.

Similarly, using word frequencies as features identifies the epilogue closely with the rest of the work. NBC and SVM classifiers attribute chapter 8 to Xenophon with a higher liklihood than some of the other chapters, both when the other \emph{Cyropaedia} chapters are used as a training text and when \emph{Anabasis} chapters are used as training texts. This information can be very useful in interpreting Xenophon's motives behind writing the \emph{Cyropaedia} and the lessons he wishes to impart. It suggests that anti-Persian sentiment was present in Xenophon's works.

More broadly, analyzing the authorship of Xenophon's \emph{Cyropaedia} provides an example for future studies of classical works. Although more analysis is needed to best determine how to avoid content dependency and accurately represent an Ancient Greek text, modern techniques like SVM can be applied to Ancient Greek. This study also demonstrates how authorship attribution techniques can be modified to perform anomaly detection. Many works have been lost over the past millennia, and if a text is of disputed authorship, it is very possible that no work written by the real author survives for comparison. As in the case of the \emph{Cyropaedia}, it can be impossible to determine even the century during which the text may have been modified, so finding a set of candidate authors becomes a hopeless task. The use of modern authorship attribution to perform anomaly detection on classical texts can greatly improve our understanding of prior literature, which in turn allows us to better understand prior civilizations.

\bstctlcite{bstctl:etal, bstctl:nodash, bstctl:simpurl}
\bibliographystyle{IEEEtranS}
\bibliography{references}

\begin{thebibliography}{10}
\providecommand{\url}[1]{#1}
\csname url@samestyle\endcsname
\providecommand{\newblock}{\relax}
\providecommand{\bibinfo}[2]{#2}
\providecommand{\BIBentrySTDinterwordspacing}{\spaceskip=0pt\relax}
\providecommand{\BIBentryALTinterwordstretchfactor}{4}
\providecommand{\BIBentryALTinterwordspacing}{\spaceskip=\fontdimen2\font plus
\BIBentryALTinterwordstretchfactor\fontdimen3\font minus
  \fontdimen4\font\relax}
\providecommand{\BIBforeignlanguage}[2]{{%
\expandafter\ifx\csname l@#1\endcsname\relax
\typeout{** WARNING: IEEEtranS.bst: No hyphenation pattern has been}%
\typeout{** loaded for the language `#1'. Using the pattern for}%
\typeout{** the default language instead.}%
\else
\language=\csname l@#1\endcsname
\fi
#2}}
\providecommand{\BIBdecl}{\relax}
\BIBdecl

\bibitem{Burrows89}
\BIBentryALTinterwordspacing
J.~F. Burrows, ``\BIBforeignlanguage{English}{'an ocean where each kind...':
  Statistical analysis and some major determinants of literary style},''
  \emph{\BIBforeignlanguage{English}{Computers and the Humanities}}, vol.~23,
  no. 4/5, pp. pp. 309--321, 1989. Available:
  \url{http://www.jstor.org/stable/30204370}
\BIBentrySTDinterwordspacing

\bibitem{ledgerRev}
\BIBentryALTinterwordspacing
T.~Corlett, \emph{\BIBforeignlanguage{English}{Journal of the Royal Statistical
  Society. Series D (The Statistician)}}, vol.~40, no.~1, pp. pp. 117--119,
  1991. Available: \url{http://www.jstor.org/stable/2348241}
\BIBentrySTDinterwordspacing

\bibitem{Diederich03}
J.~Diederich \emph{et~al.}, ``Authorship attribution with support vector
  machines,'' \emph{APPLIED INTELLIGENCE}, vol.~19, pp. 109--123, 2003.

\bibitem{domingos1996beyond}
P.~Domingos and M.~Pazzani, ``Beyond independence: Conditions for the
  optimality of the simple bayesian classi er,'' vol.~29, 1997, p. 103–130.

\bibitem{Gamon04}
M.~Gamon, ``Linguistic correlates of style: authorship classification with deep
  linguistic analysis features,'' \emph{In Proceedings of the 20th Annual
  Conference on Computational Linguistics}, pp. 611--617, 2004.

\bibitem{Gera}
D.~L. Gera, \emph{Xenophon's Cyropaedia: Style, Genre, and Literary
  Technique}.\hskip 1em plus 0.5em minus 0.4em\relax Oxford: Clarendon, 1993.

\bibitem{hirsch85}
\BIBentryALTinterwordspacing
S.~Hirsch, \emph{The Friendship of the Barbarians: Xenophon and the Persian
  Empire}.\hskip 1em plus 0.5em minus 0.4em\relax Tufts University, 1985.
  Available: \url{http://books.google.com/books?id=2--ZQgAACAAJ}
\BIBentrySTDinterwordspacing

\bibitem{holmes94}
D.~Holmes, ``Authorship attribution,'' \emph{Computers and the Humanities},
  vol.~28, no.~2, pp. 87--106, 1994.

\bibitem{Joa97}
T.~Joachims, ``Text categorization with support vector machines: Learning with
  many relevant features,'' \emph{In Proceedings of the 10th European Annual
  Conference on Machine Learning}, pp. 137--142, 1998.

\bibitem{Juola08}
P.~Juola, ``Authorship attribution,'' \emph{Foundations and Trends in
  Information Retrieval}, vol.~1, no.~3, pp. 233--334, 2008.

\bibitem{ledger89}
G.~Ledger, \emph{Re-counting Plato: a computer analysis of Plato's
  style}.\hskip 1em plus 0.5em minus 0.4em\relax Clarendon Press, 1989.

\bibitem{levison68}
M.~Levison, A.~Morton, and A.~Winspear, ``I.—the seventh letter of plato,''
  \emph{Mind}, vol.~77, no. 307, pp. 309--325, 1968.

\bibitem{morton}
A.~Morton, ``Authorship of greek prose,'' \emph{Journal of the Royal
  Statistical Society}, vol. 128, no.~2, pp. 169--233, 1965.

\bibitem{wallace64}
F.~Mosteller and D.~L. Wallace, ``Inference and disputed authorship: The
  federalist.''\hskip 1em plus 0.5em minus 0.4em\relax Addison-Wesley, 1964.

\bibitem{scikit-learn}
F.~Pedregosa \emph{et~al.}, ``Scikit-learn: Machine learning in {P}ython,''
  \emph{Journal of Machine Learning Research}, vol.~12, pp. 2825--2830, 2011.

\bibitem{Peng04}
F.~Peng, ``Augmenting naive bayes classifiers with statistical language
  models,'' \emph{Computer Science Department Faculty Publication Series,
  University of Massachusetts-Amherst}, 2004.

\bibitem{perseus}
\BIBentryALTinterwordspacing
(2013, Dec.) The perseus digital library. Available:
  \url{http://www.perseus.tufts.edu/hopper/}
\BIBentrySTDinterwordspacing

\bibitem{singhal2001modern}
A.~Singhal, ``Modern information retrieval: A brief overview,'' 2001.

\bibitem{Smith}
\BIBentryALTinterwordspacing
M.~Smith, ``\BIBforeignlanguage{English}{An investigation of morton's method to
  distinguish elizabethan playwrights},''
  \emph{\BIBforeignlanguage{English}{Computers and the Humanities}}, vol.~19,
  no.~1, pp. 3--21, 1985. Available: \url{http://dx.doi.org/10.1007/BF02259614}
\BIBentrySTDinterwordspacing

\bibitem{Stama09}
E.~Stamatatos, ``A survey of modern authorship attribution methods,''
  \emph{Journal of the American Society for Information Science and
  Technology}, vol.~60, pp. 538--556, 2009.

\bibitem{Wu04}
C.-J.~L. Ting-Fan~Wu and R.~C. Weng, ``Probability estimates for multi-class
  classification by pairwise coupling,'' \emph{Journal of Machine Learning},
  vol.~5, pp. 975--1005, 2004.

\bibitem{usher82}
S.~Usher and D.~Najock, ``A statistical study of authorship in the corpus
  lysiacum,'' \emph{Computers and the Humanities}, vol.~16, no.~2, pp. 85--105,
  1982.

\bibitem{Keselj03}
N.~C. C.~T. Vlado~Keselj, Fuchun~Peng, ``N-gram based author profiles for
  authorship attribution,'' \emph{Pacific Association for Computational
  Linguistics}, pp. 255--264, 2003.

\bibitem{Wencis77}
\BIBentryALTinterwordspacing
L.~Wencis, ``\BIBforeignlanguage{English}{Hypopsia and the structure of
  xenophon's anabasis},'' \emph{\BIBforeignlanguage{English}{The Classical
  Journal}}, vol.~73, no.~1, pp. pp. 44--49, 1977. Available:
  \url{http://www.jstor.org/stable/3296954}
\BIBentrySTDinterwordspacing

\end{thebibliography}
\end{document}